\documentclass{article}
\usepackage{spconf,amsmath,graphicx}
\usepackage{amsfonts}
\usepackage[dvipsnames]{xcolor}
\usepackage{hyperref}
\hypersetup{
  colorlinks,
  citecolor=blue,
  linkcolor=red,
  urlcolor=blue}

\newcommand{\vomega}{\boldsymbol{\mathbf{\omega}}}

\newcommand{\ie}{\textit{i}.\textit{e}.}

\usepackage{booktabs}
\usepackage{multirow}
\usepackage{stmaryrd}
\title{A Study of Deep Perceptual Metrics for Image quality Assessment}
\name{Rémi Kazmierczak$^{1}$, Gianni Franchi$^{1}$, Nacim Belkhir$^{2}$, Antoine Manzanera$^{1}$, David Filliat$^{1}$}
\address{$^{1}$U2IS ENSTA Paris, Institut Polytechnique de Paris, $^{2}$Safrantech, Safran Group}

\begin{document}
\maketitle
\begin{abstract}
Several metrics exist to quantify the similarity between images, but they are inefficient when it comes to measure the similarity of highly distorted images.
In this work, we propose to empirically investigate perceptual metrics based on deep neural networks for tackling the Image Quality Assessment (IQA) task. We study 
deep perceptual metrics according to different hyperparameters like the network's architecture or training procedure. Finally, we propose our multi-resolution perceptual metric (MR-Perceptual), that allows us to aggregate perceptual information at different resolutions and outperforms standard perceptual metrics on IQA tasks with varying image deformations. Our code is available at \url{https://github.com/ENSTA-U2IS/MR_perceptual}
\end{abstract}

\section{Introduction}\label{sec:intro}
Image Quality Assessment (IQA) plays an essential role in image-based applications~\cite{zhai2020perceptual, kamble2015no} where the acquisition systems or algorithms can introduce image quality variations. 
Although IQA is a well-known problem~\cite{wang2011applications},
it is difficult to define a  metric directly linked to human perception. Indeed, for humans, IQA is intuitive and effortless~\cite{pianykh2018modeling}.
Still, it remains a subjective measurement that is insufficient for validating algorithms or acquisition systems.

Several perceptual metrics have been investigated~\cite{ kamble2015no, zhai2020perceptual} for IQA, such as the 
$L_2$ Euclidean
distance, or SSIM~\cite{wang2004image}.
Yet, the human perception of image similarity relies on psychological vision mechanisms that are in a large extent unknown, then hard to implement. On the other hand, existing metrics only rely on estimating global or local variations between images. 

Deep Learning-based perceptual metrics have been first proposed in~\cite{johnson2016perceptual}, and~\cite{gatys2016neural} for the style transfer problem. It 
was
followed by several applications, for the quality of super-resolution algorithms~\cite{johnson2016perceptual}, semantic segmentation~\cite{luc2016semantic} task, or Generative Adversarial Network (GAN)~\cite{isola2017image} outputs quality. A perceptual metric is typically a 
$L_2$ distance
between features extracted from Deep Neural Networks (DNNs) after a forward pass of the input images. While some perceptual metrics like the Fréchet Inception Distance (FID)~\cite{heusel2017gans} %
are widely used to evaluate the quality of images generated by GANs, 
they are limited to the comparison of the estimated distribution of two set of images.

In this paper, we propose investigating IQA using Deep Learning-based perceptual metrics to compute the similarity between two images. Unlike previous studies~\cite{zhang2018unreasonable, talebi2018learned} %
that learn an image quality metric using DNN, we 
evaluate different DNNs and their associated hyperparameters (loss function, normalisation, resolution of input images, and the features extraction strategy),
with the main goal of identifying a 
deep perceptual distance as 
general-purpose metric closer to human perception. 

Our 
contribution is threefold:
first, we empirically investigate different DNN perceptual metrics 
related to
the network architecture. Next, we perform an ablation study highlighting the relationship between the training procedure of DNN parameters and 
the performances of deep perceptual metrics. 
Finally, we propose a perceptual metric that achieves the state-of-the-art results in unsupervised IQA by studying various hyperparameters impacting the computation of perceptual losses.

\section{Multi Resolution Perceptual metric}\label{sec:MR-Perceptua}

\subsection{Notations and formalism}\label{subsec:Notations} 

We denote 
$\mathcal{D} =\{x_i\}_{i=1}^{n}$
a dataset composed of 
$n$ images.
 Let $f_{\vomega}(x_i)$ denote the 
 output
 of the DNN $f$ with 
 trainable parameters 
 $\vomega=\{\omega_k\}$
 applied on image $x_i$.
  
  DNNs can be decomposed into 
  $B$ blocks applied sequentially. For example, we can decompose AlexNet~\cite{NIPS2012_c399862d}, which comprises five convolutional layers, into $B = 5$ blocks, where each block is a convolutional layer. Let us denote $\{ f^b_{\vomega}(x_i)\}_{b=1}^B$ the set of the feature maps of the $B$ blocks 
  output from
  image  $x_i$. For $b \in \llbracket 1,B \rrbracket $, the feature map $f^b_{\vomega}(x_i) \in \mathbb{R}^{H_{b} \times W_{b} \times C_{b}} $ is a three dimension tensor where $H_{b}$ and $ W_{b}$ represent the height and width of the feature map and $C_b$ is the number of channels.
We denote $f^b_{\vomega}(x_i)[h,w,c] $ with $(h,w,c) \in \llbracket 1,H_{b} \rrbracket \times \llbracket 1,W_{b} \rrbracket \times \llbracket 1,C_{b} \rrbracket $ a 
 
 particular value of the feature map.

\subsection{Perceptual metric}\label{subsec:Perceptual}

The process of computing a perceptual metric can be divided into three stages: %
the \textit{deep feature extraction} strategy from an image given a DNN architecture,
the \textit{normalization strategy} of the feature space, followed by the \textit{dissimilarity measure}
to compare the features. 
We now present these different stages.

\subsubsection{Deep Feature Extraction}
The \textit{deep feature extraction} is the initial step that allows representing the data 
into a
new feature space. 
Contrary to handcrafted features like GLCM~\cite{haralick1973textural}, we can use a trained  DNN to extract features at different levels of the network. 

If in some cases DNNs parameters are obtained from pretrained general purpose networks like ImageNet~\cite{simon2016imagenet}, some DNNs are finetuned to achieve better performances in a dataset tailored to the evaluation of perceptual metrics \cite{zhang2018unreasonable}

Also, in previous works, features were extracted from the image at the original dimension ($\times 1$). However, we also explore the feature results after upscaling by two the image, thanks to a bilinear interpolation ($\times 2$).
Let us denote 
$\phi(x_{i})$
the latent representation of the 
image $x_i$.

Given extracted features at different levels of a DNN, a straightforward strategy, termed as \textbf{\textit{linear features}}, takes all the 
feature maps
containing the perceptual and contextual information at different resolutions, and concatenate them.
It
is defined as follows:
\begin{equation}
 \phi_1(x_i) = \left[ f^1_{\vomega}(x_i),\ldots,f^B_{\vomega}(x_i) \right]
\end{equation}

An alternative strategy consists in combining features, using the Gram matrix
as proposed in~\cite{gatys2016neural}, allowing to extract new features termed as \textbf{\textit{quadratic features}}. The Gram matrix $G_{\vomega}^{b}$ of the layer $b$ is a square matrix of size 
$C_b\times C_b$. Let $(c_1 ,c_2) \in \llbracket 1,C_b \rrbracket ^2$
. The Gram matrix's coefficient at position $(c_1 ,c_2)$ defined for 
a
feature map $ f^b_{\vomega}(x_i)$ is given by:
$
G_{\vomega}^{b}[c_1 ,c_2](x_i)=\sum_{h, w} f^b_{\vomega}(x_i)[h,w,c_1] f^b_{\vomega}(x_i)[h,w,c_2].
$
We can 
now
define the \textbf{\textit{quadratic features}} of an image $x_i$ as:
\begin{equation}
 \phi_2(x_i) = \left[ G_{\vomega}^{1}(x_i),\ldots,G_{\vomega}^{B}(x_i) \right]
\end{equation}
On the one hand, the \textbf{\textit{linear features}} are directly linked to the content 
(layout)
of an image and 
to
the first 
moments of the feature maps.
On the other hand, \textbf{\textit{quadratic features}} are linked to the style of an image~\cite{gatys2016neural}, and capture 
stationary
information related to the second moment, \ie\ the 
covariance.

\subsubsection{Features Normalization}

Because the values 
vary in magnitudes between feature maps, it is essential to normalize them to homogenize all the layers and their importance. This work compares two normalization strategies. 
Current solutions consider 
an $L_2$
normalization, 
that divides each value by the $L_2$ norm of the feature map.
Yet it is also possible to normalize with 
$L_1$
or with a sigmoid function. We propose to normalize using a sigmoid function, bounding all values of the feature map 
within
$[0, 1]$.
Another operation which can be done is to use the ReLu function before normalising.

\subsubsection{Dissimilarity measure}

In order to quantify the difference between the latent representation of two images, 
we need
to define a dissimilarity measure $D$
in the feature space, not necessarily
limited to distance metrics. 
We expect that $D$ associated with the extracted features is linked  to human perceptual dissimilarity metric.

Typically perceptual loss uses the 
$L_2$
norm (MSE) between features. In addition, we propose to use different dissimilarities such as the 
$L_1$
norm (MAE), and the binary cross-entropy (CE).

\subsection{MR-Perceptual loss}
\label{subsec:MRPerceptual}

Classically the perceptual loss \cite{zhang2018unreasonable}  is composed of VGG \cite{simonyan2015deep}  linear features, followed by a $L_2$ normalization and the dissimilarity metrics is the MSE. In the rest of the paper we will refer to this loss as the classical perceptual loss.

\begin{figure}
    \centering
    \includegraphics[width=\columnwidth]{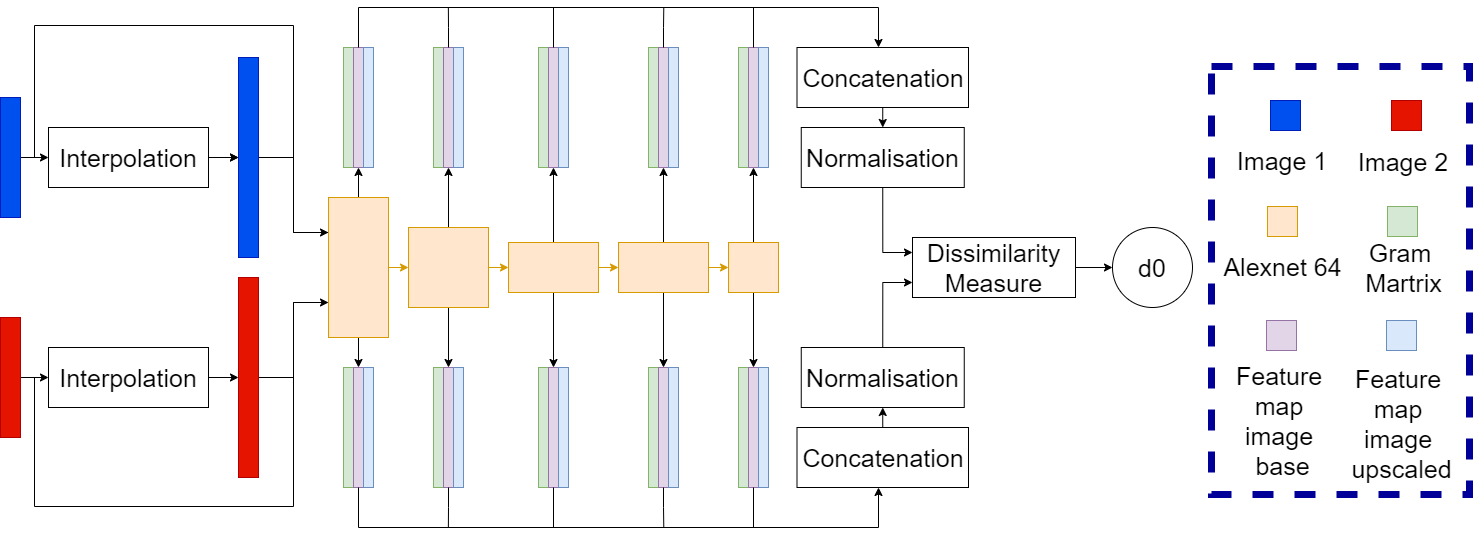}
    \caption{
    MR-Perceptual loss
    pipeline: 
    from two images of size $3 \times 64 \times 64$, we create 6 outputs (3 per image). This permits to extract more information than classical perceptual loss.}
    \label{fig:mr_perceptual_loss}
\end{figure}

Based on these three main stages, we propose to change this classical perceptual loss \cite{zhang2018unreasonable} by first proposing a multi-scale and multi-statistic feature space. Our feature space is multi-scale because instead of extracting the feature at just one resolution, we 
extract
the descriptor 
at two resolutions ($\times 1$ and $\times 2$).
Our descriptor is also multi-statistic since we concatenate quadratic and linear features for the standard resolution. We use the sigmoid function as a normalization function and then use a Binary cross-entropy as a dissimilarity measure.
The full process is illustrated in Fig. \ref{fig:mr_perceptual_loss}.

\section{Experiments}
\label{sec:format}
In this section, we first introduce the 
used
dataset 
(Sec.~\ref{subsec:Dataset}), then 
compare different feature spaces from different architectures
(Sec.~\ref{subsec:IQA_classif}). 
Next,
we discuss the importance of how to learn the representation
(Sec.~\ref{subsec:IQA_training}), and 
finally
we perform an ablation study with our technique
(Sec.~\ref{subsec:IQA_loss}).

\subsection{Dataset}
\label{subsec:Dataset}
To evaluate the performances of perceptual metrics, we use the Two-Alternative Forced-Choices (2AFC) dataset~\cite{zhang2018unreasonable}. The test sets include 36.3$K$ 
triplets
composed of one reference image 
and
two distorted images associated with their scores in $[0, 1]$ defining the ground truth perceptual dissimilarity. 
According to a human panel, the score reflects the proportion of votes for the chosen image for each tuple. 
Specifically, a ground truth will have a score of 0 if all the testers wave chosen the first image and 1 if all the testers wave chosen the second image.

The dataset is organized into six groups according to the transformations applied to the distorted images as follows: 
\textit{\textbf{Trad}} 
uses photometric and geometric transformations.\newline
\textit{\textbf{CNN}} 
uses transformations coming from DNN, like denoising autoencoders.\newline
\textit{\textbf{SuperRes}}
uses super resolution algorithms on images coming from the NITRE 2017 challenge~\cite{agustsson2017ntire}.\newline
\textit{\textbf{Deblur}} uses image extracted from video clips~\cite{su2017deep}, with video deblurring algorithms.\newline
\textit{\textbf{Color}} uses the output of image translation algorithms for image colorization applied on  ImageNet~\cite{russakovsky2015imagenet}.
\newline
\textit{\textbf{FrameInterp}} uses  different frame interpolation algorithms applied on the Davis Middlebury dataset~\cite{scharstein2002taxonomy}.

\subsection{Link between the IQA and the architectures}
\label{subsec:IQA_classif}
To evaluate the deep feature from different networks, we extracted  linear feature from AlexNet~\cite{NIPS2012_c399862d}, SqueezeNet~\cite{iandola2016squeezenet}, VGG~\cite{simonyan2015deep}, Resnet~\cite{he2016deep} and VIT~\cite{dosovitskiy2021image}. All theses networks are pretrained on imagenet \cite{russakovsky2015imagenet}.

These architectures are organized in convolutional blocks followed by dense layers. Similarly to~\cite{zhang2018unreasonable},we use the output of the five convolution blocks to extract features, except for SqueezeNet of which we use seven blocks.

Table \ref{table:result1} presents the results of a handcrafted technique (SSIM) and classical perceptual losses on different DNN architectures. The first result is that perceptual losses highly outperform SSIM, which 
supports the hypothesis of a better representation of human similarity perception.
Moreover, the AlexNet feature space seems to perform better than other architectures. Our 
interpretation
is twofold: first, ImageNet accuracy is not necessarily linked to the quality of the feature space since the tasks are different; secondly, the deeper an architecture is, the worse it performs. 
This might be linked 
to
the propagation 
of
the perceptual information throughout all the layers, such that all layers are initially trained for the ImageNet task.

\begin{table*}[h!]

\setlength{\abovecaptionskip}{0.cm}
\begin{center}

\resizebox{1.5\columnwidth}{!}
{
\begin{tabular}{lrrrrrrrrrr}
\midrule
\textbf{Loss}             & \textbf{MSE}    & \textbf{CE}        & \textbf{MSE}      & \textbf{MSE}      & \textbf{MSE}     &\textbf{MSE}       & \textbf{MSE}         &\textbf{CE}                  & \textbf{CE}         \\
\midrule
\textbf{Normalization} & \textbf{L2}       & \textbf{sigmoid} & \textbf{L2}      & \textbf{L2}    & \textbf{sigmoid} & \textbf{sigmoid}  & \textbf{L2}       &\textbf{ReLu + L1}            & \textbf{sigmoid}           \\
\midrule
\textbf{Feature}          & \textbf{Linear} &\textbf{Linear}  &\textbf{Linear} & \textbf{Linear} &\textbf{Linear}  & \textbf{Linear}  & \textbf{Quadratic} &\textbf{Linear +Quadratic} &\textbf{Linear +Quadratic} \\
\midrule
\textbf{resolution}       & $\times 1$             & $\times 1$             & $\times 2$             & $\times 1$ + $\times 2$       & $\times 1$               & $\times 1$             & $\times 2$              & $\times 1$ + $\times 2$                                & $\times 1$ + $\times 2$                 \\
\midrule
\textbf{traditional}      & 70,56                               & 71,3                                 & 72,93                               & 72,81                               & 71,74                                                               & 72,31                                & 73,02                                  & 72,66                                          & \textbf{73,78}                                 \\
\textbf{cnn}              & 83,17                               & 83,4                        & 83,07                               & 83,26                               & 83,03                                & 82,9                                 & 82,85                                  & \textbf{83,8}                                          & 83,76                                          \\
\textbf{super resolution} & 71,65                               & 71,42                                & 71,56                               & 71,7                                & 71,55                                & 71,18                                & 71,59                                  & 71,72                                          & \textbf{71,73}                                 \\
\textbf{debluring}        & 60,68                               & 61,27                                & 60,64                               & 60,8                                & 60,67                                & 60,48                                & 60,99                                  & \textbf{61,48}                                 & 61,37                                          \\
\textbf{colorization}     & \textbf{65,01}                      & 64,76                                & 63,59                               & 64,81                               & 64,78                                & 63,06                                & 64,73                                  & 64,93                                          & 64,81                                          \\
\textbf{frameinterp}      & 62,65                               & 63,55                                & 62,02                               & 62,49                               & 62,56                               & 62                                   & 62,08                                  & \textbf{63,82}                                 & 63,46                                          \\
\textbf{AVERAGE}          & 68,95                               & 69,28                                & 68,97                               & 69,31                               & 69,06                                & 68,66                                & 69,21                                  & 69,74                                          & \textbf{69,82}                  \\ \bottomrule              
\end{tabular}
}
\end{center}
\caption{\textbf{Ablation study on  AlexNet~\cite{NIPS2012_c399862d} pretrained on ImageNet with a supervised strategy}. 
}
\label{table:result3}
\end{table*}

\begin{table}
\setlength{\abovecaptionskip}{0.cm}
\Large
\begin{center}
\resizebox{\columnwidth}{!}
 {
\begin{tabular}{@{}lcccccccc@{}}
\toprule
\textbf{Datasets}&\textbf{SSIM}&\textbf{Alexnet}&\textbf{VGG}&\textbf{SqueezeNet}&\textbf{Resnet18}&\textbf{Resnet50}&\textbf{Resnet101}&\textbf{VIT}\\\midrule
\textit{\textbf{Trad}}&62,73 &70,56&70,05&\textbf{73,3}&69,66&70,73&70,71&57,68\\
\textit{\textbf{CNN}}&77,6 &\textbf{83,17}&81,28&82,64&81,59&81,43&80,88&80,38\\
\textit{\textbf{SuperRes}}&63,13 &\textbf{71,65}&69,02&70,15&69,69&68,99&68,65&64,94\\
\textit{\textbf{Deblur}}& 54,23&\textbf{60,68}&59,05&60,13&59,8&58,9&58,9&58,93\\
\textit{\textbf{Color}}& 60,89&\textbf{65,01}&60,19&63,57&60,49&60,12&59,46&63,23\\
\textit{\textbf{FrameInterp}}&57,11 &\textbf{62,65}&62,11&61,98&62,54&61,33&61,93&56,09\\
\midrule
\textbf{AVERAGE}&62,61&\textbf{68,95}&66,95&68,63&67,30&66,92&66,76&63,54\\
\textbf{\begin{tabular}[c]{@{}l@{}}
ImageNet Top1 acc
\end{tabular}}&NA&63,30&74,50&57.50&73.19&77.15&80.9&77.91\\\bottomrule
\end{tabular}
}
\end{center}
\caption{\textbf{Comparative results  of different DNN architectures \textbf{\textit{linear features}}}.The firsts rows denotes the 2AFC score. The last row shows the 
Top1 accuracy on ImageNet~\cite{russakovsky2015imagenet}.}
\label{table:result1}
\end{table}

\subsection{Link between the IQA and the training procedure}
\label{subsec:IQA_training}

In Section \ref{subsec:IQA_classif}, first experiments focused on the impact of the architecture against a classical perceptual metric. Now, we focus our experiments on the strategy to train a DNN for an optimal representation for perceptual queries. 
For this purpose,
we consider a ResNet50 architecture trained on ImageNet and  the following training strategies: a supervised training,  DeepCluster~\cite{caron2019deep}, %
Dino~\cite{caron2021emerging}, MoCo v2~\cite{chen2020improved}, OBoW~\cite{gidaris2021obow},  SimCLR~\cite{chen2020simple}, 
SwAV~\cite{caron2021unsupervised}, and 
finally
a random initialization of the parameters.
In Table \ref{table:result2}, we compare the performance of ResNet50 with the different pretrained parameters, and we observe that our supervised training outperforms 
the others in 
most of the tasks. 
This shows that
a supervised procedure helps to inject 
in the network useful information
for the perceptual task.

\begin{table}[h!]
\setlength{\abovecaptionskip}{0.cm}
\begin{center}
\resizebox{\columnwidth}{!}
 {
\begin{tabular}{@{}lcccccccccc@{}}
\toprule
Dataset & \textbf{Random} & \textbf{Supervised} & \textbf{Deepcluster} & \textbf{Dino} & \textbf{MoCo} & \textbf{Obow} & \textbf{SimCLR} & \textbf{SwAV} \\ \midrule
\textit{\textbf{Trad}}        & 58,54                                                & \textbf{70,73}                                           & 68,73                                                     & 69,78                                              & 67,91                                              & 69,74                                              & 68,97                                                & 68,77                                              \\
\textit{\textbf{CNN}}         & 80,07                                                & \textbf{81,43}                                           & 80,21                                                     & 80,03                                              & 78,3                                               & 79,04                                              & 79,12                                                & 79,74                                              \\
\textit{\textbf{SuperRes}}    & 65,97                                                & \textbf{68,99}                                           & 66,7                                                      & 66,25                                              & 67,17                                              & 65,88                                              & 67,43                                                & 66,48                                              \\
\textit{\textbf{Deblur}}      & 59,32                                                & \textbf{58,9}                                            & 58,26                                                     & 58,13                                              & 58,45                                              & 57,83                                              & 57,92                                                & 58,09                                              \\
\textit{\textbf{Color}}       & \textbf{63,03}                                       & 60,12                                                    & 55,92                                                     & 56,12                                              & 56,17                                              & 55,81                                              & 56,37                                                & 56,26                                              \\
\textit{\textbf{FrameInterp}} & 56,99                                                & 61,33                                                    & 61,94                                                     & 62,27                                              & 62,45                                              & 61,54                                              & 61,66                                                & \textbf{62,48}                                     \\
\textit{\textbf{AVERAGE}}     & 63,99                                                & \textbf{66,92}                                           & 65,29                                                     & 65,43                                              & 65,08                                              & 64,97                                              & 65,25                                                & 65,3\\ \bottomrule                                              
\end{tabular}
}
\end{center}
\caption{\textbf{Comparative results 
showing the impact of supervised training
on 2AFC~\cite{zhang2018unreasonable} with Resnet 50~\cite{he2016deep} architecture}. We run a linear feature extraction 
with different pretraining conditions.}
\label{table:result2}
\end{table}

\subsection{Importance of the different components for IQA}

\label{subsec:IQA_loss}
Based on previous results in Sections \ref{subsec:IQA_classif} and \ref{subsec:IQA_training}, AlexNet trained on ImageNet in a supervised manner is the best to quantify the perceptual dissimilarity. We studied in the table \ref{table:result5} the performance according to the extracted features and observed that features extracted from  4$^{th}$ and 5$^{th}$ layers 
are the best for the \textbf{\textit{Trad}} set. But features extracted from the 2$^{nd}$ and 3$^{rd}$ 
outperform on \textbf{\textit{CNN}}, \textbf{\textit{SuperRes}}, \textbf{\textit{Deblur}} and \textbf{\textit{Color}} distortions. 

This suggests that some layers might focus on particular details in the distorted images which provides clues for being invariant to some distortions.

\begin{table}[h!]
\setlength{\abovecaptionskip}{0.cm}
\begin{center}
\resizebox{\columnwidth}{!}
 {
\begin{tabular}{@{}lcccccccccc@{}}
\toprule
Dataset & \textbf{Block 1} & \textbf{Block 2} & \textbf{Block 3} & \textbf{Block 4} & \textbf{Block 5} & \textbf{All}\\ \midrule
\textit{\textbf{Trad}} & 59,15 & 69,35 & 71,97 & 72,92 & \textbf{73,29} & 70,56 \\
\textit{\textbf{CNN}} & 81,65 & 82,88 & \textbf{82,94} & 82,84 & 82,03 & 83,17 \\
\textit{\textbf{SuperRes}} & 65,94 & 71,6 & \textbf{71,63} & 71,19 & 70,6 & 71,65 \\
\textit{\textbf{Deblur}} & 59,08 & \textbf{60,87} & 60,67 & 60,43 & 60,1 & 60,68  \\
\textit{\textbf{Color}} & 62,82 & \textbf{64,73} & 64,42 & 63,88 & 63,9 & 65,01 \\
\textit{\textbf{FrameInterp}} & 57,23 & 61,95 & 62,69 & 62,67 & \textbf{62,71} & 62,65  \\
\textbf{AVERAGE} & 64,31 & 68,56 & 69,05 & 68,99 & 68,77 & 68,95\\ \bottomrule
\end{tabular}
}
\end{center}
\caption{\textbf{Comparative results 
showing the impact the chosen layer
on 2AFC~\cite{zhang2018unreasonable} with AlexNet ~\cite{NIPS2012_c399862d} architecture}. The \textbf{bolded} results shows the best results among blocks}
\label{table:result5}
\end{table}

Table \ref{table:result3} shows an ablation study to evaluate the
relevant
hyperparameters 
for designing a novel perceptual metric as detailed in Section \ref{subsec:Perceptual}. We 
consider
the features extraction strategies, the type of dissimilarity metric in the feature space, the normalization strategy, and finally, the resolution. Multi-resolution seems to be the key to improving performances. In addition, Multi statistic seems to improve also the performances for certain distortions.

As shown in 
Tab.~\ref{table:result4},
the classic perceptual metric setup is outperformed in all the distortions; this 
remains
true for all the networks, including Watching \cite{DBLP:journals/corr/PathakGDDH16}, Split-brain \cite{zhang2017splitbrain}, Puzzle \cite{noroozi2017unsupervised} and BiGAN \cite{DBLP:journals/corr/DonahueKD16} .

\begin{table}%
\setlength{\abovecaptionskip}{0.cm}
\begin{center}
\resizebox{\columnwidth}{!}
{
\begin{tabular}{@{}lcccccccc@{}}
\toprule
Dataset & \textbf{Ours} & \textbf{Watching} & \textbf{Split-Brain} & \textbf{Puzzle} & \textbf{BiGAN} \\ \midrule
\textit{\textbf{Trad}} &  \textbf{73,8} & 66,5 & 69,5 & 71,5 & 69,8 \\
\textit{\textbf{CNN}} & \textbf{83,8} & 80,7 & 81,4 & 82,0 & 83,0 \\
\textit{\textbf{SuperRes}} & \textbf{71.7} & 69,6 & 69,6 & 70,2 & 70,7 \\
\textit{\textbf{Deblur}} & \textbf{61,4} & 60,6 & 59.3 & 60,2 & 60,5 \\
\textit{\textbf{Color}} & \textbf{64,9} & 64,4 & 64,3 & 62,8 & 63,7 \\
\textit{\textbf{FrameInterp}} & \textbf{63.5} & 61,6 & 61,1 & 61,8 & 62,5\\
\textbf{AVERAGE} & \textbf{69,8} & 67,2 & 67,5 & 68,1 & 68,4\\  
\bottomrule
\end{tabular}
}
\end{center}
\caption{\textbf{Comparison of our Alexnet with MR-perceptual loss with the setup presented in \cite{zhang2018unreasonable}}.}
\label{table:result4}
\end{table}

\section{Conclusions}
We empirically investigated general-purpose deep perceptual metrics w.r.t. different experimental settings on an IQA task. First, we show that it is unnecessary to use deeper DNN with complex architecture; a simple AlexNet is sufficient for perceptual metrics. Despite convincing results of self-supervised, we show that a supervised strategy remains the best choice. Finally, we confirm that combining features at different resolutions is relevant as it forces 
the DNN to be more robust against various types of distortion.
Future work would involve combining this new perceptual metric and image-to-image translation DNNs to improve the quality of 
generated images.

\bibliographystyle{IEEEbib}
\bibliography{strings,refs}

\begin{thebibliography}{10}

\bibitem{zhai2020perceptual}
Guangtao Zhai and Xiongkuo Min,
\newblock ``Perceptual image quality assessment: a survey,''
\newblock {\em Science China Information Sciences}, vol. 63, no. 11, pp.
  211301, 2020.

\bibitem{kamble2015no}
Vipin Kamble and KM~Bhurchandi,
\newblock ``No-reference image quality assessment algorithms: A survey,''
\newblock {\em Optik}, vol. 126, no. 11-12, pp. 1090--1097, 2015.

\bibitem{wang2011applications}
Zhou Wang,
\newblock ``Applications of objective image quality assessment methods
  [applications corner],''
\newblock {\em IEEE signal processing magazine}, vol. 28, no. 6, pp. 137--142,
  2011.

\bibitem{pianykh2018modeling}
Oleg~S Pianykh, Ksenia Pospelova, and Nick~H Kamboj,
\newblock ``Modeling human perception of image quality,''
\newblock {\em Journal of digital imaging}, vol. 31, no. 6, pp. 768--775, 2018.

\bibitem{wang2004image}
Zhou Wang, Alan~C Bovik, Hamid~R Sheikh, and Eero~P Simoncelli,
\newblock ``Image quality assessment: from error visibility to structural
  similarity,''
\newblock {\em IEEE transactions on image processing}, vol. 13, no. 4, pp.
  600--612, 2004.

\bibitem{johnson2016perceptual}
Justin Johnson, Alexandre Alahi, and Li~Fei-Fei,
\newblock ``Perceptual losses for real-time style transfer and
  super-resolution,''
\newblock in {\em ECCV}. Springer, 2016, pp. 694--711.

\bibitem{gatys2016neural}
LA~Gatys, AS~Ecker, and M~Bethge,
\newblock ``A neural algorithm of artistic style,''
\newblock in {\em 16th Annual Meeting of the Vision Sciences Society (VSS
  2016)}. Scholar One, Inc., 2016, p. 326.

\bibitem{luc2016semantic}
Pauline Luc, Camille Couprie, Soumith Chintala, and Jakob Verbeek,
\newblock ``Semantic segmentation using adversarial networks,''
\newblock in {\em NIPS Workshop on Adversarial Training}, 2016.

\bibitem{isola2017image}
Phillip Isola, Jun-Yan Zhu, Tinghui Zhou, and Alexei~A Efros,
\newblock ``Image-to-image translation with conditional adversarial networks,''
\newblock in {\em CVPR}, 2017, pp. 1125--1134.

\bibitem{heusel2017gans}
Martin Heusel, Hubert Ramsauer, Thomas Unterthiner, Bernhard Nessler, and Sepp
  Hochreiter,
\newblock ``Gans trained by a two time-scale update rule converge to a local
  nash equilibrium,''
\newblock {\em Advances in neural information processing systems}, vol. 30,
  2017.

\bibitem{zhang2018unreasonable}
Richard Zhang, Phillip Isola, Alexei~A Efros, Eli Shechtman, and Oliver Wang,
\newblock ``The unreasonable effectiveness of deep features as a perceptual
  metric,''
\newblock in {\em CVPR}, 2018, pp. 586--595.

\bibitem{talebi2018learned}
Hossein Talebi and Peyman Milanfar,
\newblock ``Learned perceptual image enhancement,''
\newblock in {\em 2018 IEEE international conference on computational
  photography (ICCP)}. IEEE, 2018, pp. 1--13.

\bibitem{NIPS2012_c399862d}
Alex Krizhevsky, Ilya Sutskever, and Geoffrey~E Hinton,
\newblock ``Imagenet classification with deep convolutional neural networks,''
\newblock in {\em Advances in Neural Information Processing Systems},
  F.~Pereira, C.~J.~C. Burges, L.~Bottou, and K.~Q. Weinberger, Eds. 2012,
  vol.~25, Curran Associates, Inc.

\bibitem{haralick1973textural}
Robert~M Haralick, Karthikeyan Shanmugam, and Its'~Hak Dinstein,
\newblock ``Textural features for image classification,''
\newblock {\em IEEE Transactions on systems, man, and cybernetics}, , no. 6,
  pp. 610--621, 1973.

\bibitem{simon2016imagenet}
Marcel Simon, Erik Rodner, and Joachim Denzler,
\newblock ``Imagenet pre-trained models with batch normalization,''
\newblock {\em arXiv preprint arXiv:1612.01452}, 2016.

\bibitem{simonyan2015deep}
Karen Simonyan and Andrew Zisserman,
\newblock ``Very deep convolutional networks for large-scale image
  recognition,''
\newblock in {\em ICLR}, 2015.

\bibitem{agustsson2017ntire}
Eirikur Agustsson and Radu Timofte,
\newblock ``Ntire 2017 challenge on single image super-resolution: Dataset and
  study,''
\newblock in {\em CVPR workshops}, 2017, pp. 126--135.

\bibitem{su2017deep}
Shuochen Su, Mauricio Delbracio, Jue Wang, Guillermo Sapiro, Wolfgang Heidrich,
  and Oliver Wang,
\newblock ``Deep video deblurring for hand-held cameras,''
\newblock in {\em CVPR}, 2017, pp. 1279--1288.

\bibitem{russakovsky2015imagenet}
Olga Russakovsky, Jia Deng, Hao Su, Jonathan Krause, Sanjeev Satheesh, Sean Ma,
  Zhiheng Huang, Andrej Karpathy, Aditya Khosla, Michael Bernstein, et~al.,
\newblock ``Imagenet large scale visual recognition challenge,''
\newblock {\em International journal of computer vision}, vol. 115, no. 3, pp.
  211--252, 2015.

\bibitem{scharstein2002taxonomy}
Daniel Scharstein and Richard Szeliski,
\newblock ``A taxonomy and evaluation of dense two-frame stereo correspondence
  algorithms,''
\newblock {\em International journal of computer vision}, vol. 47, no. 1, pp.
  7--42, 2002.

\bibitem{iandola2016squeezenet}
Forrest~N. Iandola, Song Han, Matthew~W. Moskewicz, Khalid Ashraf, William~J.
  Dally, and Kurt Keutzer,
\newblock ``Squeezenet: Alexnet-level accuracy with 50x fewer parameters and
  <0.5mb model size,'' 2016.

\bibitem{he2016deep}
Kaiming He, Xiangyu Zhang, Shaoqing Ren, and Jian Sun,
\newblock ``Deep residual learning for image recognition,''
\newblock in {\em CVPR}, 2016, pp. 770--778.

\bibitem{dosovitskiy2021image}
Alexey Dosovitskiy, Lucas Beyer, Alexander Kolesnikov, Dirk Weissenborn,
  Xiaohua Zhai, Thomas Unterthiner, Mostafa Dehghani, Matthias Minderer, Georg
  Heigold, Sylvain Gelly, Jakob Uszkoreit, and Neil Houlsby,
\newblock ``An image is worth 16x16 words: Transformers for image recognition
  at scale,'' 2021.

\bibitem{caron2019deep}
Mathilde Caron, Piotr Bojanowski, Armand Joulin, and Matthijs Douze,
\newblock ``Deep clustering for unsupervised learning of visual features,''
  2019.

\bibitem{caron2021emerging}
Mathilde Caron, Hugo Touvron, Ishan Misra, Hervé Jégou, Julien Mairal, Piotr
  Bojanowski, and Armand Joulin,
\newblock ``Emerging properties in self-supervised vision transformers,'' 2021.

\bibitem{chen2020improved}
Xinlei Chen, Haoqi Fan, Ross Girshick, and Kaiming He,
\newblock ``Improved baselines with momentum contrastive learning,'' 2020.

\bibitem{gidaris2021obow}
Spyros Gidaris, Andrei Bursuc, Gilles Puy, Nikos Komodakis, Matthieu Cord, and
  Patrick P{\'e}rez,
\newblock ``Obow: Online bag-of-visual-words generation for self-supervised
  learning,''
\newblock in {\em 2021 CVPR}. IEEE, 2021, pp. 6826--6836.

\bibitem{chen2020simple}
Ting Chen, Simon Kornblith, Mohammad Norouzi, and Geoffrey Hinton,
\newblock ``A simple framework for contrastive learning of visual
  representations,'' 2020.

\bibitem{caron2021unsupervised}
Mathilde Caron, Ishan Misra, Julien Mairal, Priya Goyal, Piotr Bojanowski, and
  Armand Joulin,
\newblock ``Unsupervised learning of visual features by contrasting cluster
  assignments,'' 2021.

\bibitem{DBLP:journals/corr/PathakGDDH16}
Deepak Pathak, Ross~B. Girshick, Piotr Doll{\'{a}}r, Trevor Darrell, and
  Bharath Hariharan,
\newblock ``Learning features by watching objects move,''
\newblock {\em CoRR}, vol. abs/1612.06370, 2016.

\bibitem{zhang2017splitbrain}
Richard Zhang, Phillip Isola, and Alexei~A. Efros,
\newblock ``Split-brain autoencoders: Unsupervised learning by cross-channel
  prediction,'' 2017.

\bibitem{noroozi2017unsupervised}
Mehdi Noroozi and Paolo Favaro,
\newblock ``Unsupervised learning of visual representations by solving jigsaw
  puzzles,'' 2017.

\bibitem{DBLP:journals/corr/DonahueKD16}
Jeff Donahue, Philipp Kr{\"{a}}henb{\"{u}}hl, and Trevor Darrell,
\newblock ``Adversarial feature learning,''
\newblock {\em CoRR}, vol. abs/1605.09782, 2016.

\end{thebibliography}

\end{document}